\documentclass{article}

\usepackage{arxiv}

\usepackage[utf8]{inputenc} 
\usepackage[T1]{fontenc}    
\usepackage{hyperref}       
\usepackage{url}            
\usepackage{booktabs}       
\usepackage{amsfonts}       
\usepackage{nicefrac}       
\usepackage{microtype}      
\usepackage{lipsum}		
\usepackage{graphicx}
\usepackage{natbib}
\usepackage{doi}

\usepackage{graphicx}%
\usepackage{multirow}%
\usepackage{amsmath,amssymb,amsfonts}%
\usepackage{amsthm}%
\usepackage{mathrsfs}%
\usepackage[title]{appendix}%
\usepackage{xcolor}%
\usepackage{textcomp}%
\usepackage{manyfoot}%
\usepackage{booktabs}%
\usepackage{algorithm}%
\usepackage{algorithmicx}%
\usepackage{algpseudocode}%
\usepackage{listings}%
\usepackage[table]{xcolor}
\usepackage{csquotes}
\usepackage{tikz}

\usepackage{listings}   
\usepackage{xcolor}     

\definecolor{backgroundgray}{RGB}{245,245,245}

\title{Reflecting in the Reflection: Integrating a Socratic Questioning Framework into Automated AI-Based Question Generation}

\author{{Ondřej Holub} \\
	Department of Applied Mathematics\\
	Czech Technical University in Prague\\
	Prague, Czechia\\
	\texttt{holubon6@cvut.cz} \\
	\And
	{Essi Ryymin} \\
	Unit for Research and Development of Higher Education Pedagogy\\
	Häme University of Applied Sciences\\
	Hämeenlinna, Finland\\
	\texttt{essi.ryymin@hamk.fi} \\
\And
	{Rodrigo Alves} \\
	Department of Applied Mathematics\\
	Czech Technical University in Prague\\
	Prague, Czechia\\
	\texttt{rodrigo.alves@fit.cvut.cz} \\
}

\renewcommand{\shorttitle}{Integrating a Socratic Questioning Framework into Automated AI-Based Question Generation}

\hypersetup{
pdftitle={Integrating a Socratic Questioning Framework into Automated AI-Based Question Generation},
pdfsubject={q-bio.NC, q-bio.QM},
pdfauthor={Ondřej Holub,  Essi Ryymin, Rodrigo Alves},
pdfkeywords={Large language model, automatic question generation, Socratic method, reflection questions, reflection in education},
}

\begin{document}
\maketitle

\begin{abstract}
	Designing good reflection questions is pedagogically important but time-consuming and unevenly supported across teachers. This paper introduces a \emph{reflection-in-reflection} framework for automated generation of reflection questions with large language models (LLMs). Our approach coordinates two role-specialized agents, a \emph{Student--Teacher} and a \emph{Teacher--Educator}, that engage in a Socratic multi-turn dialogue to iteratively refine a single question given a teacher-specified topic, key concepts, student level, and optional instructional materials. The Student--Teacher proposes candidate questions with brief rationales, while the Teacher--Educator evaluates them along clarity, depth, relevance, engagement, and conceptual interconnections, responding only with targeted coaching questions or a fixed signal to stop the dialogue. 
We evaluate the framework in an authentic lower-secondary ICT setting on the topic, using GPT-4o-mini as the backbone model and a stronger GPT-4–class LLM as an external evaluator in pairwise comparisons of clarity, relevance, depth, and overall quality. First, we study how interaction design and context (dynamic vs.\ fixed iteration counts; presence or absence of student level and materials) affect question quality. Dynamic stopping combined with contextual information consistently outperforms fixed 5- or 10-step refinement, with very long dialogues prone to drift or over-complication. Second, we show that our two-agent protocol produces questions that are judged substantially more relevant and deeper, and better overall, than a one-shot baseline using the same backbone model.
\end{abstract}



\section{Introduction}
\label{sec:introduction}

Reflection questions are prompts or queries designed to encourage learners (or sometimes educators) to critically think about and examine their own thinking, processes, decisions, assumptions, and outcomes in relation to scientific inquiry, experimentation, or learning (e.g., what they learned, what they did, why they did it, how they might do it differently)~\cite{sijmkens2023scaffolding,zarestky2022reflective}. Moreover, studies shows that reflection questions foster metacognition (thinking about one’s thinking) and can strengthen conceptual understanding, awareness of one’s methods, and recognition of underlying assumptions or biases~\cite{white1998inquiry,ankita2024metacognitive}.

Yet producing prompts that truly elicit deep reflection is demanding work: they must align with concrete learning outcomes, calibrate cognitive demand (beyond recall toward analysis, synthesis, and application), and be inclusive of diverse backgrounds and ways of expressing understanding~\cite{menekse2022different}. In practice, several challenges arise. For example, (a) instructors, especially those teaching large or multi-section courses, often have limited time to design and refine such prompts~\cite{dyment2010quality,luo2016determining}; and (b) the level of expertise in generating reflection questions may differ considerably among members of the teaching team. As a result, many courses rely on generic or recycled questions whose clarity, relevance, and challenge level are uneven. This variability tends to yield superficial responses, provides limited diagnostic value for instructors, and can widen equity gaps when prompts presume specific prior knowledge or norms. The practical bottleneck is clear: designing high-quality reflection prompts at scale is time-intensive and inconsistent, constraining students’ opportunities for meaningful reflection and instructors’ ability to use reflections for formative feedback~\cite{luo2016determining}.

To mitigate these constraints and enable more specialized prompts within limited instructor time, researchers have widely explored the problem of automatic question generation (AQG), using a variety of methods for the question generation (QG) process~\cite{LeiteCardoso2023,mulla2023automatic}. AQG programmatically produces candidate questions conditioned on course materials, learning outcomes, and cognitive taxonomies, offering a scalable way to diversify and iterate on prompt design. Modern approaches span rule- or template-based methods, retrieval-augmented pipelines, and neural generators that can target specific skills (e.g., analysis or synthesis) and adapt to varied learner profiles. In educational settings, AQG can reduce authoring effort, broaden the range of prompt types, and support personalization; beyond the classroom, it is useful for self-testing and interview preparation. 

However, most prior AQG systems~\cite{kurdi2020systematic} have focused on generating multiple-choice, true/false, or fill-in-the-blank items that primarily assess factual recall. These systems generally do not design questions to support reflective practice, despite its importance in learning. The few efforts that target open-ended prompts are typically unguided, which often leads to superficial responses and increases the risk of biased or construct-misaligned questions. Finally, prior work~\cite{das2021automatic} is often validated  with limited ablation studies and inadequate reproducibility reporting, which undermines generalizability and fair comparisons across systems.

\begin{figure}
  \centering
  \begin{tikzpicture}[
      box/.style={draw, rectangle},
      bigbox/.style={draw, rectangle, minimum width=8cm, minimum height=4cm},
      arrow/.style={->, >=stealth, thick}
    ]
    \node[box] (ht) at (0,0) {Human Teacher};

    \node[bigbox] (system) at (7.5,0) {};
    \node[above] at (system.north) {Question Generation System};

    \node[box, align=center] (st) at (5.5,0) {Student--Teacher\\ \footnotesize (AI Agent)};
    \node[box, align=center] (te) at (9.5,0) {Teacher--Educator\\ \footnotesize (AI Agent)};

    \draw[arrow] (te.south) to[bend left=30] node[below] {Guiding Questions} (st.south |- te.south);
    \draw[arrow] (st.north |- te.north) to[bend left=30] node[above] {Generated Question} (te.north);

    \draw[arrow] (ht.north) to[bend left=30] node[above] {Topic \& Concepts} (system.west |- ht.north);
    \draw[arrow] (system.west |- ht.south) to[bend left=30] node[below] {Reflection Question} (ht.south);
  \end{tikzpicture}
  \caption[Reflection-in-Reflection AQG Architecture]{Reflection-in-reflection question-generation architecture. A Human Teacher provides topic \& concepts (and optional materials) to the Question Generation System. Inside the system, a \emph{Student–Teacher} LLM drafts reflection questions while a \emph{Teacher–Educator} LLM applies a Socratic framework to pose guiding questions; their iterative loop refines the draft. The system returns the finalized reflection question to the instructor.}

  \label{fig:system-architecture}
\end{figure}
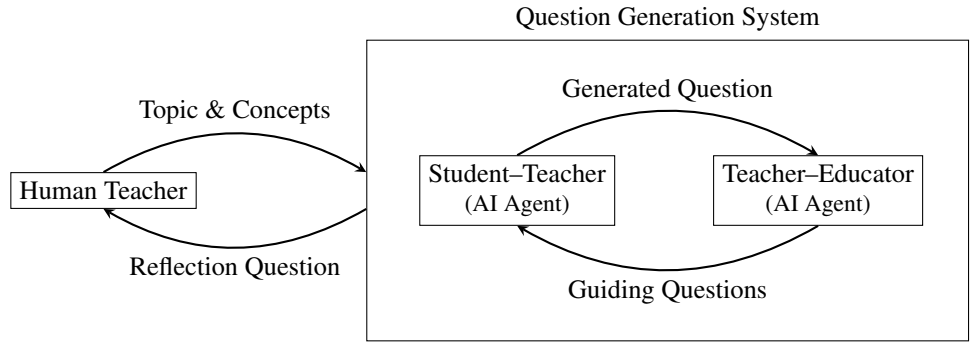

In this paper, we address this research gap by proposing a \emph{reflection-in-reflection} framework for the automated generation of reflection questions using generative artificial intelligence (AI) models, grounded in educational theory. The scheme of our framework (illustrated in Figure~\ref{fig:system-architecture}) can be briefly summarized as follows: (1) A \textbf{Human Teacher} specifies the \emph{topic and core concepts}. At this stage, the teacher may also provide supporting materials related to the topic (e.g., instructional resources in digital form). This information is (2) directed to the \emph{Question Generation System}, an AI system that (3) processes the inputs and returns a reflection question to the teacher. Rather than prompting a single generative AI model directly, our approach employs two independent AI agents, the \textbf{Student–Teacher} and the \textbf{Teacher–Educator}. By aiming for breadth and creativity while remaining on topic, the first acts as a novice instructor, drafting \emph{reflection} questions conditioned on the human instructor’s inputs. The second agent, the \emph{Teacher–Educator}, serves as a pedagogical coach; instead of rewriting the prompt, it applies a \emph{Socratic questioning framework} to pose \emph{guiding questions} that probe clarity, alignment with learning outcomes, cognitive demand, relevance, accessibility, and fairness. The Student–Teacher and Teacher–Educator agents iterate on these drafts and guiding questions until a satisfactory outcome is reached. Finally, (4) the system returns the finalized \emph{reflection question} to the human teacher, who may accept it as is, edit it, or provide new constraints to trigger another cycle.

Note that, both the \emph{Student–Teacher} and \emph{Teacher–Educator} are LLM-based agents with distinct role prompts. Their interaction mirrors an effective mentoring exchange: the Student–Teacher proposes, while the Teacher–Educator interrogates and scaffolds through Socratic moves, leading to iterative improvement. This design aims to (i) elevate questions beyond factual recall toward analysis, synthesis, and application; (ii) surface assumptions and potential biases; and (iii) reduce instructor authoring time while preserving pedagogical intent. In essence, our approach embeds an \emph{educational reflection framework} (operationalized through \emph{Socratic questioning}) into the AI generation process, enabling the system to reflect on its own outputs while producing reflection-oriented questions for learners. 

Our main contributions can be summarized as follows:

\begin{itemize}
    \item We propose a \emph{reflection-in-reflection} framework grounded in \emph{Socratic questioning} for the automated generation of reflection questions to be used in learning environments. The framework integrates educational theory with generative AI to produce reflective prompts that align with learning outcomes and foster deeper metacognitive engagement.

    \item We validate and evaluate our approach using an large language model (LLM)-based evaluator, providing empirical evidence of the framework’s effectiveness and usefulness. Our experiments are conducted on real instructional materials and authentic learning scenarios, offering practical insights and guidance for future research and system development.

    \item We perform an ablation study on the diverse inputs of our method to examine the individual contribution of each component, highlighting their respective benefits and the overall robustness of the framework.

    \item We provide \emph{actionable guidance to educators} on how to leverage LLMs for designing effective reflection prompts. The framework translates core pedagogical requirements (e.g, clarity, alignment with learning outcomes, appropriate cognitive demand) into concrete prompting strategies that practitioners can adopt to author deeper and more equitable reflective tasks.

\end{itemize}

The remainder of this paper is organized as follows. Section~\ref{sec:literature-review} reviews the relevant literature. Section~\ref{sec:methodology} describes the proposed system and its underlying methodology. Section~\ref{sec:experiments} presents the experimental setup and discusses the results. Finally, Section~\ref{sec:conclusion} concludes the paper and outlines directions for future work.

\section{Literature Review}
\label{sec:literature-review}

\subsection{Reflection}
Reflection is a critical component of the learning process, with its importance recognized across various fields, including psychology, sociology, and philosophy~\cite{clara2015reflection}{damen2017reflection}. It involves complex cognitive, emotional, and social processes that encourage deeper understanding and self-assessment~\cite{burke2023if}. Roots of reflection trace back to ancient Greece and the Socratic method of questioning~\cite{openstax2022socrates}. Modern educational theories, notably from Dewey and Schön, have further established reflection as a cornerstone of experiential learning and professional development~\cite{dewey1910, schon1983reflective}. Dewey defined reflective thought as the \enquote{active, persistent and careful consideration of any belief or supposed form of knowledge in the light of the grounds that support it and the further conclusion to which it tends}~\cite{dewey1910}. Schön later introduced the concepts of \textit{reflection-in-action} (reflecting during an event) and \textit{reflection-on-action} (reflecting after an event)~\cite{schon1983reflective}.

From a cognitive perspective, reflection is closely intertwined with metacognition, the monitoring and regulation of one’s own thinking. Metacognitive models emphasize two core components: knowledge about cognition and regulation of cognition~\cite{schraw1994assessing}. Reflection prompts, when well-designed, function as regulatory supports that guide learners to evaluate strategies, identify assumptions, monitor progress, and articulate reasoning. This connection is reinforced by research showing that reflective activities can strengthen self-regulated learning, deepen conceptual understanding, and support equitable participation in inquiry-based environments~\cite{white1998inquiry}.

However, creating effective reflection prompts remains challenging. Reflection quality depends heavily on the structure, clarity, and cognitive level of the prompt~\cite{dyment2010quality,menekse2022different}. Poorly designed prompts can lead to superficial responses, exacerbate cognitive load, or unintentionally privilege students with stronger prior knowledge. This practical difficulty motivates the need for scalable methods for designing pedagogically robust reflective questions.

For the purpose of this work, reflection is defined according to Moon as \enquote{a mental process with purpose and/or outcome in which manipulation of meaning is applied to relatively complicated or unstructured ideas in learning or to problems for which there is no obvious solution}~\cite[p.~161]{moon1999reflection}. Consequently, \textit{reflection questions} are prompts designed to facilitate this process, guiding learners to analyze their experiences and build a new understanding.

\subsection{Educational Models for Reflection}
To structure and guide reflection, various educational models have been developed. Among the most influential is the Socratic method, a form of cooperative dialogue that uses systematic, disciplined questioning to stimulate critical thinking and self-discovery~\cite{paul2006socratic}. In this method, a teacher acts as a facilitator, guiding a student through a series of questions to examine their beliefs and assumptions. The goal is not to provide answers directly but to help the student arrive at their own understanding~\cite{openstax2022socrates}. This dialogic process of inquiry and refinement closely mirrors the process of reflection, making it a foundational model for this work's approach to question generation.

Socratic questions typically target clarification, evidence, implications, viewpoints, or alternatives, cognitive moves essential to reflective practice~\cite{king1992facilitating}. Recent psychological analyses emphasize that Socratic dialogue promotes cognitive conflict and self-examination, which aligns naturally with metacognitive learning goals~\cite{van2017reflection,clara2015reflection}. Other frameworks also provide a pedagogical basis for designing effective reflection prompts. Bloom's Taxonomy, for instance, categorizes educational objectives into levels of cognitive complexity. Reflection aligns with its higher-order thinking skills: analyzing, evaluating, and creating, which require learners to move beyond simple recall and engage in deeper cognitive processing~\cite{anderson2001taxonomy}. Similarly, models such as Gibbs' Reflective Cycle offer a structured sequence for analyzing experiences through stages like description, feelings, evaluation, analysis, conclusion, and an action plan~\cite{gibbs1988learning}. While this work is primarily inspired by the Socratic method, these models reinforce the importance of structured inquiry in fostering meaningful reflection.

\subsection{Automated Question Generation}
The goal of AQG systems is to automatically create questions that are relevant, meaningful, and appropriate for the intended audience. Traditional rule-based systems use predefined linguistic patterns to transform declarative sentences into questions. While interpretable, these systems are often difficult to scale and lack diversity~\cite{LeiteCardoso2023, heilman2010}.

More recent neural network-based approaches, particularly those using transformer architectures and large language models (LLMs), treat QG as a sequence generation task. These models, often fine-tuned on large datasets, can generate more fluent and diverse questions~\cite{mulla2023automatic, Lopez2021}. While many AQG systems focus on factual, multiple-choice, or fill-in-the-blank questions, this work addresses the gap in generating high-quality reflection questions. By leveraging the Socratic method within a collaborative LLM-based system, this research aims to produce questions that promote deeper, more meaningful reflection.

\section{Methodology}
\label{sec:methodology}

This section details the design choices behind our \emph{reflection-in-reflection} framework, which coordinates two purpose-specific LLM  agents: the \textbf{Student--Teacher} and the \textbf{Teacher--Educator}. The aim is to co-produce high-quality reflection questions under the supervision of a human teacher (see Figure~\ref{fig:system-architecture}). The method emphasizes (1) role separation to reduce cognitive overload and drift, (2) iterative coaching rather than one-shot rewriting, and (3) light constraints that maintain direction without limiting creativity. All decisions were taken with classroom use in mind: transparency of the process, adaptability to different subjects and levels, and \emph{modest computational} cost.

\subsection{Dialogue Dynamics}
\label{subsec:dynamics}

The human teacher initiates the process by specifying a topic $T$ and, optionally, a target student level $L$ and a finite set of key concepts $C$ relevant to that topic. The teacher may also attach supplementary digital resources (e.g., PDF lecture notes), which we denote collectively as $M$. Together, $(T, C, L, M)$ define the initial context shared by both agents.

\begin{algorithm}[t]
\caption{Reflection-in-Reflection Generation Loop}
\label{alg:reflection-in-reflection}
\begin{algorithmic}[1]
    \Require Topic $T$, concept set $C$, student level $L$, optional materials $M$
    \Ensure Final reflection question $q^\star$ and dialogue trace $\mathcal{D}$

    \State $\mathcal{D} \gets \emptyset$ \Comment{initialize dialogue trace}

    \State \textbf{(Initialization)}
    \State Provide $(T, C, L, M)$ to the Student--Teacher agent
    \State Prompt and receive the initial reflection question $q_0$ and brief rationale $r_0$
    \State $\mathcal{D} \gets \mathcal{D} \cup \{(q_0, r_0)\}$

     \State ~
    \State \textbf{(Iterative refinement)}
    \For{$k = 1, 2, \dots$}
        \State \textbf{(Teacher--Educator)}
        \State Provide $(T, C, L, M, q_{k-1})$ to the Teacher--Educator agent
        \State Receive a single Socratic coaching question $s_k$
        \State $\mathcal{D} \gets \mathcal{D} \cup \{s_k\}$
        \State ~
        \State \textbf{(Student--Teacher)}
        \State Provide $(T, C, L, M, q_{k-1}, s_k)$ to the Student--Teacher agent
        \State Receive revised question $q_k$ and brief explanation of changes $r_k$
        \State ~
        \State $\mathcal{D} \gets \mathcal{D} \cup \{(q_k, r_k)\}$
        \State ~

        \State \textbf{(Teacher--Educator)}
        \State Query Teacher--Educator with $q_k$
        \If{Teacher--Educator returns the fixed token \texttt{Great question!}}
            \State $q^\star \gets q_k$
            \State \textbf{break}
        \EndIf
    \EndFor

    \State \Return $q^\star$, $\mathcal{D}$
\end{algorithmic}
\end{algorithm}

Operationally, the dialogue can be viewed as a three-phase iterative procedure, which is formally summarized in Algorithm~\ref{alg:reflection-in-reflection}. In the initialization phase, the Student--Teacher produces a single reflection question $q_0$ that is tailored to the given level $L$ and, when appropriate, grounded in the supplementary materials $M$. In each refinement phase, the Teacher--Educator examines the current question $q_{k-1}$ and, instead of rewriting it, poses one Socratic coaching question $s_k$ targeting specific pedagogical properties (e.g., clarity or depth). In the revision phase, the Student--Teacher updates the question to $q_k$ in light of $s_k$ and briefly explains the modifications. These three phases are repeated until the Teacher--Educator explicitly signals that the question meets the desired criteria or a convergence criteria (e.g., a fixed number of iterations) is reached. The result is a single, polished reflection question that has been iteratively improved in a controlled yet flexible manner.

\textbf{REMARK:} From a design perspective, our aim is to rely on relatively lightweight, general-purpose LLMs that can, in principle, be deployed locally on modest hardware or by using accessible generative AI APIs, rather than assuming access to large models and GPU infrastructure or \emph{high-end} APIs. This makes the approach more accessible across diverse educational settings, including institutions with limited technical or financial resources. To compensate for the reduced capacity of such models, much of the pedagogical sophistication is encoded in the \emph{interaction protocol} rather than in the model itself: the refinement process is explicitly structured as an implementation of a well-known educational reflection framework based on Socratic questioning. This is interesting from a research perspective because it shifts part of the “intelligence” from model size to algorithmic design, allowing us to systematically study how role separation, turn-by-turn guidance, and theoretically grounded question types influence the quality of generated prompts.

Note that, generally, the iterative loop is preferable to a one-shot generation process because open-ended reflection questions must simultaneously satisfy multiple criteria (e.g., clarity, depth, relevance). In a single-step setting, these criteria are implicitly balanced within a single model call, which often leads to prompts that are either too shallow or unnecessarily complex. The iterative scheme separates these concerns across turns: the Teacher--Educator agent can focus on one salient issue at a time, and the Student--Teacher agent can make targeted, interpretable revisions. We observed that, this structure mirrors human instructional feedback cycles and makes the evolution of each question traceable and explainable to educators, while still leaving sufficient room for creative variation in the generated prompts.

\subsection{Student--Teacher Agent}
\label{subsec:Student--Teacher}

Conditioned on the context $(T, C, L, M)$, we instantiate a \emph{Student–Teacher} agent, implemented as an LLM configured to act as a learner who is training to design reflection questions under the supervision of the \emph{Teacher–Educator}. Rather than producing a set of independent questions, the Student–Teacher maintains and iteratively improves a \emph{single} reflection question through a sequence of revision steps guided by feedback. This design explicitly mirrors an apprenticeship setting: the Student–Teacher proposes an initial question, receives targeted critique, and then revises its work in light of that critique.

The interaction with the Student–Teacher agent is realized through two complementary prompts. In the \emph{initial generation prompt}, the agent is instructed to generate one reflection question based on the topic $T$, the set of key concepts $C$, the  target student level $L$, and, when present, the supplementary materials $M$ (e.g., PDF lecture notes). After formulating the question, the Student–Teacher is required to provide a brief rationale, constrained to at most five sentences, explaining its design choices and how the question connects to the supplied context and materials. This ``question + rationale'' format encourages the agent to make its reasoning explicit and grounded in $T$, $C$, $L$, and $M$, rather than relying on unfocused or purely associative generation. Note that it also ensures that the revision history is transparent and easy to audit.

In subsequent turns, we employ a \emph{revision prompt} that explicitly situates the Student–Teacher as a learner responding to the Teacher–Educator’s feedback. The prompt states that the Teacher–Educator has commented on the previous question and rationale iteration, and instructs the Student–Teacher to (i) analyze this feedback and (ii) revise the question accordingly. The agent is reminded that the revised question must be appropriate for the specified level $L$ and that it should \emph{not} attempt to weave all available concepts from $C$ into a single prompt; instead, it should focus on one concept or a small subset, to avoid unnecessary complexity and cognitive overload. The Student–Teacher is again required to return only the revised question together with a short explanation (maximum five sentences) that justifies the changes in light of the Teacher–Educator’s comments.

Taken together, the initial and revision prompts implement a constrained Socratic dialogue in which the Student–Teacher functions as a novice reflecting on its own question design, and the Teacher–Educator acts as a mentor probing alignment, clarity, and cognitive demand. The single-question, feedback-driven setup emphasizes depth of refinement rather than breadth of generation, while the requirement to continuously explain revisions provides a traceable record of the agent’s evolving reasoning and its responsiveness to pedagogical guidance.

\subsection{Teacher--Educator Agent}
\label{subsec:Teacher--Educator}

The Teacher--Educator is also implemented as an LLM, but configured to act as a pedagogical coach rather than a content generator. Conditioned on the same context $(T, C, L, M)$ as the Student--Teacher, it receives the current candidate reflection question together with the Student--Teacher’s brief explanation. Its role is to diagnose strengths and weaknesses of this proposal and to pose a \emph{single} Socratic question that will most productively guide the Student--Teacher’s next revision step. By restricting the Teacher--Educator to one question per turn and prohibiting direct rewrites, the design preserves the Student--Teacher’s ownership of the question while yielding focused, high-signal feedback. This constrained role also reduces the risk of topic drift, unnecessary wholesale rewriting of the question, and the introduction of unsupported or extraneous content, thereby maintaining continuity with prior revisions while still promoting targeted improvement.

The feedback criteria used by the Teacher--Educator combine insights from AQG and reflection-focused pedagogy. Surveys of AQG systems and evaluation protocols commonly emphasize properties such as clarity, relevance, answerability, and appropriate difficulty or cognitive level as central to question quality, including for open-ended items \cite{mulla2023automatic,LeiteCardoso2023,Mucciaccia2025,Scaria2024}. In parallel, the reflection literature highlights the importance of prompts that elicit metacognitive engagement, connect to learners’ experiences, and support the articulation of reasoning rather than mere recall \cite{sijmkens2023scaffolding,zarestky2022reflective,white1998inquiry,ankita2024metacognitive}. Drawing on these strands, we instantiate five operational dimensions for reflection questions: (1) \emph{clarity} (linguistic and pragmatic comprehensibility), (2) \emph{depth} (orientation toward analysis, evaluation, or synthesis), (3) \emph{relevance} (alignment with $T$, $C$, $L$, and $M$), (4) \emph{engagement} (potential to invite personal connection and discussion), and (5) \emph{interconnections} (when appropriate, prompting learners to relate multiple ideas without overloading the task).

These dimensions are encoded explicitly in a single Teacher--Educator prompt. The prompt states that the agent is a Teacher Educator guiding a Student--Teacher in developing reflective questions, and it re-presents the topic $T$, the key concept set $C$, and the student level $L$, as well as indicating that the first user turn may include materials $M$ (e.g., PDF lecture notes) that the Student--Teacher can draw upon. The Teacher--Educator is instructed to analyze the Student--Teacher’s question and explanation with respect to the five dimensions above and to respond with exactly one question, introduced with the fixed prefix ``\texttt{The Teacher's feedback:}'', that will help the Student--Teacher refine the question. When it judges that the question is already of high quality, it is allowed to terminate the dialogue by returning only the phrase ``\texttt{Great question!}''.

The Socratic nature of the Teacher--Educator’s intervention is made explicit by providing canonical question stems grounded in established work on Socratic questioning and critical thinking \cite{paul2006socratic,dalim2022promoting}. The prompt lists examples such as ``What do you mean by \dots?'', ``How does this relate to \dots?'', ``Can you provide an example of \dots?'', ``What assumptions are you making?'', and ``What are the implications of \dots?'', and instructs the agent to adapt such stems to the current candidate question. Crucially, the Teacher--Educator is also reminded that the Student--Teacher ``should not needlessly connect all the concepts mentioned in the materials'', and thus should avoid overloading a single prompt with the full concept set $C$. Instead, the Teacher--Educator is expected to steer revisions toward questions that target one concept or a small subset, with appropriate depth and accessibility for level $L$.

Overall, this design yields a Teacher--Educator agent whose behavior is constrained by an explicit, theory-informed rubric and by disciplined Socratic questioning. Rather than allowing the LLM to rely on an opaque internal notion of “depth’’ or “quality’’, the prompt operationalizes these constructs through the five dimensions and example stems, producing feedback that is more stable, interpretable, and aligned with both AQG evaluation practices and educational aims for reflective learning. Taken together, the two-agent setup leverages the Student--Teacher’s capacity, inherited from large-scale pretraining on diverse corpora, to explore a broad creative space of candidate questions, while the Teacher--Educator constrains and reshapes this creativity so that the resulting prompts remain pedagogically grounded, level-appropriate, and aligned with the intended learning outcomes.

\section{Experiments}
\label{sec:experiments}

We structure our empirical study around two main research questions.

\paragraph{RQ1 (Role of Context and Interaction Design).}
How do different sources of contextual information and interaction dynamics influence the quality of generated reflection questions? Here we examine the impact of (i) explicitly specifying the target \emph{student level} $L$, (ii) providing or withholding supporting instructional \emph{materials} $M$, and (iii) using a \emph{dynamic} stopping criterion versus a fixed number of refinement iterations. We investigate how the presence or absence of these elements affects the four evaluation criteria above, and whether particular configurations systematically lead to clearer, more relevant, or deeper reflection questions.

\paragraph{RQ2 (Protocol Effectiveness).}
Does the proposed \emph{reflection-in-reflection} protocol, in which a Student--Teacher and a Teacher--Educator agent iteratively co-construct a question, produce reflection prompts that are judged superior to those generated directly by a single LLM in a one-shot setting? More concretely, we ask whether the two-agent, Socratic, multi-turn architecture yields higher \emph{clarity}, \emph{relevance}, \emph{depth}, and \emph{overall quality} than a baseline that prompts the same backbone model once using only the topic and concepts.

\subsection{Domain and Auxiliary Data}
\label{subsec:data}

To evaluate the proposed \emph{reflection-in-reflection} framework, we conducted experiments within a well-defined educational domain: ``\emph{The Fundamentals of Information Technology''}. The particular focus is on introductory concepts related to how the internet works. This domain was chosen because it is commonly taught at the lower secondary level (in Czechia) and provides a clear set of technical concepts suitable for generating structured reflective prompts.

The instructional materials used in our study were provided by an experienced teacher specializing in digital literacy and ICT education. These materials were originally designed for 8th--9th grade students (approximately ages 13--15) and represent authentic resources used in real classroom settings. Using such domain-validated materials ensured that the generated reflection questions aligned with realistic curricular goals, age-appropriate cognitive demand, and typical student prior knowledge.

The supporting materials consisted of:
\begin{itemize}
	\item \textbf{Slide-based presentations:} visually structured materials introducing key concepts such as servers, data centers, client–server communication, routers, IP addresses, and packet transmission, supported by diagrams, animations, and analogies (e.g., transport routes) to illustrate how data moves through the internet;
	\item \textbf{Teacher guidelines and notes:} detailing learning objectives, key concepts (e.g., decentralization, servers, routers, packets), step-by-step lesson flow, and classroom activities such as analogies (e.g., “dinosaurus” for packet fragmentation) and discussion prompts used to probe student understanding.
	
\end{itemize}

\noindent Together, these artifacts formed the optional resource set $M$ described in Section~\ref{sec:methodology}, which was made available to both AI agents during question generation.
From these materials, we extracted a set of core concepts relevant to the topic, which served as the concept set $C$ for the Student--Teacher and Teacher--Educator agents:

\begin{itemize}
	\item Basics of how the internet works
	\item Decentralization of the internet
	\item Servers, data centers, and routers
	\item Server--client relationships
	\item Data packets
	\item IP addresses
\end{itemize}

\subsection{Evaluation Protocol}
\label{subsec:evaluation}

Evaluating the quality of generated open-ended questions is a persistent challenge in AQG. Unlike factual or multiple-choice questions, reflective prompts often require subjective judgments (e.g., related to clarity, depth, or pedagogical usefulness), which made fully automatic evaluation difficult in pre-AI times. Traditionally, open-ended questions are assessed by human annotators, who rate them along several qualitative dimensions. Although human evaluation offers nuanced insight, particularly for reflective or conceptual items, it is costly, time-consuming, and susceptible to variability across annotators. These limitations have motivated research toward more scalable alternatives~\cite{gorgun2024exploring,deroy2024mirror}, especially given the combinatorial number of possibilities that must be analyzed.

Thus, reliance solely on human annotation is impractical for iterative system development or large-scale experiments, particularly in settings with high combinatorial task complexity like ours. Recent advances therefore incorporate LLMs as evaluators~\cite{Mucciaccia2025, Scaria2024}, allowing researchers to approximate human-like judgments while benefiting from greater consistency and scalability. These LLM-based evaluators act as surrogate annotators, applying the same criteria used in human evaluation, and prior work has demonstrated their effectiveness in capturing many of the qualitative aspects traditionally assessed by humans. In line with these developments, our evaluation adopts an LLM-based assessment framework rather than human annotation, aligning with recent studies that employ LLMs for structured qualitative evaluation of open-ended questions, combining the interpretive richness of human evaluation with the efficiency of automated methods.

More formally, we follow the common recommendation that the evaluator should be strictly more powerful than the models being assessed, in order to reduce evaluation bias and avoid overfitting to idiosyncrasies of a single backbone. Concretely, we use GPT\mbox{-}4o\mbox{-}mini as the backbone model for our agents and a larger, more capable GPT\mbox{-}4–class model as the evaluator. The latter has substantially higher capacity (in terms of architecture and parameter count; $8$B vs $1.76$T) and is thus better suited to act as an external judge, rather than merely mirroring the behaviour of the agent models it evaluates.

Guided by prior AQG literature and the specific requirements of reflective questioning~\cite{LeiteCardoso2023, Mucciaccia2025, Scaria2024}, we evaluate generated questions along four criteria: \emph{clarity}, \emph{relevance}, \emph{depth}, and \emph{overall quality}. Clarity consolidates format, grammar, and fluency into a single notion of how clearly the question is stated and how easy it is to understand. Relevance measures alignment with the target topic and key concepts supplied by the human designer. Depth captures the extent to which a question encourages critical thinking and non-trivial reflection, rather than eliciting simple factual or yes/no responses. Overall quality provides a holistic judgement of structure, engagement, and pedagogical usefulness. Each criterion is operationalized as a concise guiding question for the evaluator (e.g., “Is the question clearly stated and easy to understand?”). For full prompt formulations and implementation details, we refer the reader to our code repository, where all evaluation prompts are made available.

Evaluating hundreds of questions within a single prompt is cognitively demanding for humans and can also degrade LLM performance due to context-length limitations and attention dilution~\cite{liu2024lost,li2024loogle}. Therefore, we adopt a pairwise evaluation protocol. For each topic and associated concept set, the evaluator is presented with a pair of candidate questions and a single evaluation criterion at a time. The LLM returns a \emph{difference score} $d \in \{-2,-1,1,2\}$ indicating both the preferred question and the strength of preference, together with a brief textual justification. Negative scores indicate that the first question is better than the second, positive scores the converse, and the absence of a zero value forces a strict preference in each comparison. Aggregating these pairwise scores across topics and concept sets yields comparative performance measures for different agent configurations and criteria.

\newpage


\subsection{RQ1: Role of Context and Interaction Design}
\label{subsec:rq1-design}

To investigate how context and interaction design affect question quality, we consider three iteration regimes for the reflection-in-reflection protocol:
\begin{itemize}
    \item \textbf{DYN}: a \emph{dynamic} scheme in which the Teacher--Educator can terminate the dialogue as soon as it emits the fixed token \texttt{Great question!};
    \item \textbf{F05}: a \emph{fixed} scheme in which the dialogue always runs for exactly 5 refinement iterations, regardless of the Teacher--Educator’s feedback;
    \item \textbf{F10}: a \emph{fixed} scheme with 10 refinement iterations.
\end{itemize}
These regimes are combined with two further binary factors: (i) whether the student level is explicitly specified (L $\in \{\checkmark,\times\}$), and (ii) whether supporting materials are provided (M $\in \{\checkmark,\times\}$). Each configuration can thus be written as
\[
\alpha = (\text{IT}, \text{L}, \text{M}) \in \mathcal{C},
\]
where $\mathcal{C}$ contains $12$ combinations (matching the rows/columns in the heatmaps in Figures~\ref{HM:overall}--\ref{HM:depth}). For each configuration $\alpha \in \mathcal{C}$ and each topic, we run the generation pipeline and obtain $k$ final reflection questions, yielding a set
\[
Q_\alpha = \{q_{\alpha,1}, \dots, q_{\alpha,k}\}
\]
per configuration (aggregated across topics). To compare two configurations $\alpha$ and $\beta$, we perform pairwise LLM evaluations over all ordered pairs
\[
(q, q') \in Q_\alpha \times Q_\beta
\]
under a fixed evaluation criterion (cf.~Section~\ref{subsec:evaluation}). To mitigate position bias, for each pair we randomly decide which question appears as \emph{Question~1} and which as \emph{Question~2}. The evaluator returns a difference score
$d_{\text{raw}} \in \{-2,-1,1,2\}$,
where negative values indicate that \emph{Question 1} is preferred, positive values indicate that \emph{Question 2} is preferred, and the magnitude encodes the strength of this preference.

For the purpose of comparing configurations, we re-orient these scores so that negative values always favour $\alpha$ and positive values favour $\beta$. Concretely, for each comparison we define
\[
\tilde{d} =
\begin{cases}
-\,d_{\text{raw}}, & \text{if the question from configuration } \alpha \text{ was shown as Question 1},\\[2pt]
d_{\text{raw}}, & \text{if the question from configuration } \alpha \text{ was shown as Question 2},
\end{cases}
\]
so that $\tilde{d} > 0$ means a preference for $\alpha$ and $\tilde{d} < 0$ a preference for $\beta$. Let $D_{\alpha,\beta}$ denote the multiset of all such re-oriented scores $\tilde{d}$ when comparing configuration $\alpha$ against configuration $\beta$. We then define a normalized preference index
$\gamma(\alpha,\beta) \in [0,1]$
by, first,  mapping each $\tilde{d} \in D_{\alpha,\beta}$ to
\[
s(\tilde{d}) = \frac{2 + \tilde{d}}{4} \in \{1, 0.75, 0.25, 0\},
\]
so that a strong preference for $\alpha$ (i.e., $\tilde{d} = 2$) maps to $1$, and a strong preference for $\beta$ (i.e., $\tilde{d} = -2$) maps to $0$, with intermediate values for mild preferences. The overall index is then
\[
\gamma(\alpha,\beta)
= \frac{1}{|D_{\alpha,\beta}|} \sum_{\tilde{d} \in D_{\alpha,\beta}} s(\tilde{d}).
\]

By construction, $\gamma(\alpha,\beta) = 1$ if \emph{all} comparisons strongly favour questions from configuration $\alpha$, and $\gamma(\alpha,\beta) = 0$ if they all strongly favour $\beta$. Values near $0.5$ indicate no consistent advantage of either configuration. These $\gamma(\alpha,\beta)$ scores form the basis of the pairwise heatmaps analyzed for RQ1, summarizing how iteration regime, explicit student-level specification, and the presence of materials influence the perceived quality of generated reflection questions.

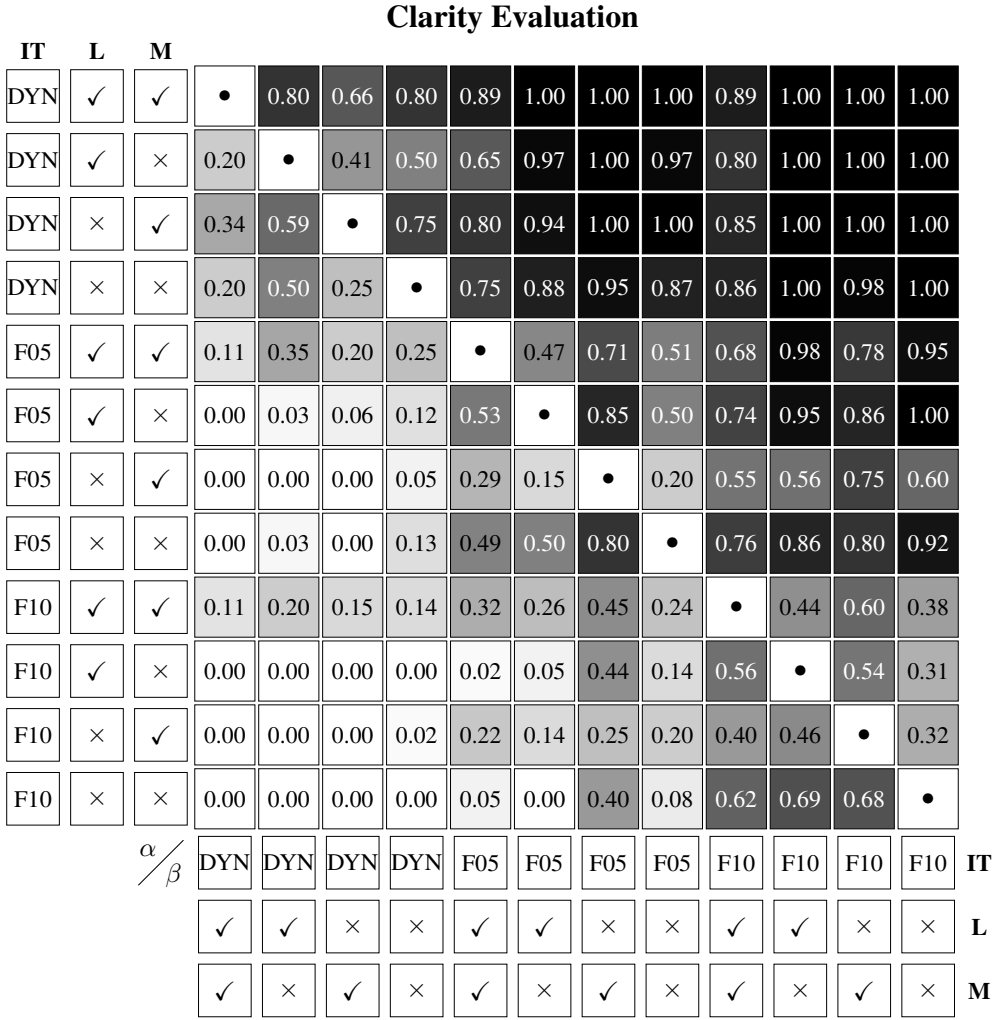
\begin{figure}
	\centering
	\begin{tikzpicture}[scale=0.85]
		
		\node[font=\large\bfseries] at (7.5,0.2) {Clarity Evaluation};
		
		\foreach \y [count=\n] in {
			{-1, 0.80, 0.66, 0.80, 0.89, 1.00, 1.00, 1.00, 0.89, 1.00, 1.00, 1.00},
            {0.20, -1, 0.41, 0.50, 0.65, 0.97, 1.00, 0.97, 0.80, 1.00, 1.00, 1.00},
            {0.34, 0.59, -1, 0.75, 0.80, 0.94, 1.00, 1.00, 0.85, 1.00, 1.00, 1.00},
            {0.20, 0.50, 0.25, -1, 0.75, 0.88, 0.95, 0.87, 0.86, 1.00, 0.98, 1.00},
            {0.11, 0.35, 0.20, 0.25, -1, 0.47, 0.71, 0.51, 0.68, 0.98, 0.78, 0.95},
            {0.00, 0.03, 0.06, 0.12, 0.53, -1, 0.85, 0.50, 0.74, 0.95, 0.86, 1.00},
            {0.00, 0.00, 0.00, 0.05, 0.29, 0.15, -1, 0.20, 0.55, 0.56, 0.75, 0.60},
            {0.00, 0.03, 0.00, 0.13, 0.49, 0.50, 0.80, -1, 0.76, 0.86, 0.80, 0.92},
            {0.11, 0.20, 0.15, 0.14, 0.32, 0.26, 0.45, 0.24, -1, 0.44, 0.60, 0.38},
            {0.00, 0.00, 0.00, 0.00, 0.02, 0.05, 0.44, 0.14, 0.56, -1, 0.54, 0.31},
            {0.00, 0.00, 0.00, 0.02, 0.22, 0.14, 0.25, 0.20, 0.40, 0.46, -1, 0.32},
            {0.00, 0.00, 0.00, 0.00, 0.05, 0.00, 0.40, 0.08, 0.62, 0.69, 0.68, -1},
		} {
			
			\foreach \x [count=\m] in \y {
				
				\ifnum\m=\n
				\node[minimum size=8mm,draw=black,line width=0.1pt,fill=white] at (\m+2,-\n) {$\bullet$}; 
				\else
				
				\pgfmathparse{\x<0.50 ? "black" : "white"}
				\edef\textcol{\pgfmathresult}
				
				\pgfmathparse{\x*100}
				\edef\mix{\pgfmathresult}
				
				\pgfmathparse{\m+2}
				\edef\newM{\pgfmathresult}
				
				\node[fill=black!\mix!white,minimum size=8mm,line width=0.1pt,draw=black,text=\textcol] at (\newM,-\n) {\footnotesize \x};
				
				\fi
			}
			
		}
		
		\node[minimum size=6mm] at (0,-0.3) {\footnotesize \textbf{IT}};
		\node[minimum size=6mm] at (1,-0.3) {\footnotesize \textbf{L}};
		\node[minimum size=6mm] at (2,-0.3) {\footnotesize \textbf{M}};

		\foreach \a [count=\i] in {DYN,DYN,DYN,DYN,F05,F05,F05,F05,F10,F10,F10,F10} {
			\node[line width=0.2pt,draw=black,inner sep=0pt,text width=7mm,minimum height=7mm,align=center] at (0,-\i) {\footnotesize \a};
		}
		
		\foreach \a [count=\i] in {$\checkmark$,$\checkmark$,$\times$,$\times$,$\checkmark$,$\checkmark$,$\times$,$\times$,$\checkmark$,$\checkmark$,$\times$,$\times$} {
			\node[minimum size=7mm,draw=black,line width=0.2pt] at (1,-\i) {\a};
		}
		
		\foreach \a [count=\i] in {$\checkmark$,$\times$,$\checkmark$,$\times$,$\checkmark$,$\times$,$\checkmark$,$\times$,$\checkmark$,$\times$,$\checkmark$,$\times$} {
			\node[minimum size=7mm,draw=black,line width=0.2pt] at (2,-\i) {\a};
		}
		
		\foreach \a [count=\i] in {DYN,DYN,DYN,DYN,F05,F05,F05,F05,F10,F10,F10,F10} {
			\pgfmathparse{\i+2}
			\edef\newI{\pgfmathresult}
			\node[line width=0.2pt,draw=black,inner sep=0pt,text width=7mm,minimum height=7mm,align=center] at (\newI,-13) {\footnotesize \a};
		}
		
		\foreach \a [count=\i] in {$\checkmark$,$\checkmark$,$\times$,$\times$,$\checkmark$,$\checkmark$,$\times$,$\times$,$\checkmark$,$\checkmark$,$\times$,$\times$} {
			\pgfmathparse{\i+2}
			\edef\newI{\pgfmathresult}
			\node[minimum size=7mm,draw=black,line width=0.2pt] at (\newI,-14) {\a};
		}

		\foreach \a [count=\i] in {$\checkmark$,$\times$,$\checkmark$,$\times$,$\checkmark$,$\times$,$\checkmark$,$\times$,$\checkmark$,$\times$,$\checkmark$,$\times$} {
			\pgfmathparse{\i+2}
			\edef\newI{\pgfmathresult}
			\node[minimum size=7mm,draw=black,line width=0.2pt] at (\newI,-15) {\a};
		}
		
		\node[minimum size=6mm] at (14.8,-13) {\footnotesize \textbf{IT}};
		\node[minimum size=6mm] at (14.8,-14) {\footnotesize \textbf{L}};
		\node[minimum size=6mm] at (14.8,-15) {\footnotesize \textbf{M}};
		
		\node[minimum size=6mm] at (1.8,-12.8) {$\alpha$};
		\node[minimum size=6mm] at (2.2,-13.2) {$\beta$};
		
		\node[minimum size=6mm,path picture={\draw (path picture bounding box.south west) -- (path picture bounding box.north east);}] at (2,-13) { };

               \node[minimum size=15mm] at (14.8,-16) {\footnotesize \textbf{}};
        \node[minimum size=35mm] at (14.8,2) {\footnotesize \textbf{}};

	\end{tikzpicture}
	\caption{Heatmap evaluation of the \textbf{Clarity} component of the generated questions. Rows represent configurations $\alpha$ and columns represent configurations $\beta$. Each entry shows the score $\gamma(\alpha,\beta)$. 
		\textbf{Legend:} \emph{IT}: Iteration Type; \emph{L}: Student Level; \emph{M}: Presence of Supporting Materials; \emph{DYN}: Dynamic number of iterations; \emph{F05}: Fixed to 5 iterations; \emph{F10}: Fixed to 10 iterations.}
	
	\label{HM:clarity}
\end{figure}

The resulting $\gamma(\alpha,\beta)$ values are visualized as pairwise heatmaps in Figures~\ref{HM:clarity}–\ref{HM:overall}, where each figure corresponds to one evaluation dimension (overall quality, clarity, relevance, and depth, respectively). Each matrix entry at row $\alpha$ and column $\beta$ shows $\gamma(\alpha,\beta)\in[0,1]$, with darker cells indicating a stronger preference for the row configuration over the column configuration, and lighter cells indicating the opposite. By construction, for any two distinct configurations $\alpha$ and $\beta$ we have
\[
\gamma(\alpha,\beta) + \gamma(\beta,\alpha) = 1.
\]
Thus, entries mirrored across the main diagonal sum to one. For example, in Figure~\ref{HM:overall} the cell at row~1, column~2 corresponds to $\gamma(\alpha,\beta)=0.03$ with $\alpha = (\texttt{DYN}, L\checkmark, M\checkmark)$ and $\beta = (\texttt{DYN}, L\checkmark, M\times)$, indicating that questions from configuration $\beta$ are almost always preferred over those from $\alpha$. The symmetric cell at row~2, column~1 then shows $\gamma(\beta,\alpha)=0.97 = 1 - 0.03$, reflecting the same comparison but with the roles of row and column reversed. The main diagonal is left blank because comparing a configuration with itself would trivially yield $\gamma(\alpha,\alpha)=0.5$ and is therefore uninformative. We generated five different questions for each possible $\alpha$ profile, meaning that each cell in Figures~\ref{HM:clarity}–\ref{HM:overall} summarizes twenty-five pairwise comparisons, resulting in a total of $3,300$ LLM-based evaluations.

\begin{figure}
	\centering
	\begin{tikzpicture}[scale=0.85]
		
		\node[font=\large\bfseries] at (7.5,0.2) {Relevance Evaluation};
		
		\foreach \y [count=\n] in {
			{-1, 0.03, 0.23, 0.11, 0.45, 0.43, 1.00, 0.39, 0.91, 0.82, 1.00, 0.75},
            {0.97, -1, 0.86, 0.77, 0.79, 0.97, 1.00, 0.72, 1.00, 1.00, 1.00, 1.00},
            {0.77, 0.14, -1, 0.54, 0.69, 0.67, 1.00, 0.58, 0.98, 0.98, 0.98, 0.92},
            {0.89, 0.23, 0.46, -1, 0.72, 0.71, 1.00, 0.69, 1.00, 1.00, 1.00, 0.95},
            {0.55, 0.21, 0.31, 0.28, -1, 0.32, 0.91, 0.31, 0.89, 0.51, 0.94, 0.50},
            {0.57, 0.03, 0.33, 0.29, 0.68, -1, 1.00, 0.57, 0.92, 0.76, 1.00, 0.82},
            {0.00, 0.00, 0.00, 0.00, 0.09, 0.00, -1, 0.04, 0.69, 0.12, 0.87, 0.04},
            {0.61, 0.28, 0.42, 0.31, 0.69, 0.43, 0.96, -1, 0.94, 0.59, 1.00, 0.66},
            {0.09, 0.00, 0.02, 0.00, 0.11, 0.08, 0.31, 0.06, -1, 0.11, 0.55, 0.10},
            {0.18, 0.00, 0.02, 0.00, 0.49, 0.24, 0.88, 0.41, 0.89, -1, 0.96, 0.41},
            {0.00, 0.00, 0.02, 0.00, 0.06, 0.00, 0.13, 0.00, 0.45, 0.04, -1, 0.04},
            {0.25, 0.00, 0.08, 0.05, 0.50, 0.18, 0.96, 0.34, 0.90, 0.59, 0.96, -1},
		} {
			
			\foreach \x [count=\m] in \y {
				
				\ifnum\m=\n
				\node[minimum size=8mm,draw=black,line width=0.1pt,fill=white] at (\m+2,-\n) {$\bullet$}; 
				\else
				
				\pgfmathparse{\x<0.50 ? "black" : "white"}
				\edef\textcol{\pgfmathresult}
				
				\pgfmathparse{\x*100}
				\edef\mix{\pgfmathresult}
				
				\pgfmathparse{\m+2}
				\edef\newM{\pgfmathresult}
				
				\node[fill=black!\mix!white,minimum size=8mm,line width=0.1pt,draw=black,text=\textcol] at (\newM,-\n) {\footnotesize \x};
				
				\fi
			}
			
		}
		
		\node[minimum size=6mm] at (0,-0.3) {\footnotesize \textbf{IT}};
		\node[minimum size=6mm] at (1,-0.3) {\footnotesize \textbf{L}};
		\node[minimum size=6mm] at (2,-0.3) {\footnotesize \textbf{M}};

		\foreach \a [count=\i] in {DYN,DYN,DYN,DYN,F05,F05,F05,F05,F10,F10,F10,F10} {
			\node[line width=0.2pt,draw=black,inner sep=0pt,text width=7mm,minimum height=7mm,align=center] at (0,-\i) {\footnotesize \a};
		}
		
		\foreach \a [count=\i] in {$\checkmark$,$\checkmark$,$\times$,$\times$,$\checkmark$,$\checkmark$,$\times$,$\times$,$\checkmark$,$\checkmark$,$\times$,$\times$} {
			\node[minimum size=7mm,draw=black,line width=0.2pt] at (1,-\i) {\a};
		}
		
		\foreach \a [count=\i] in {$\checkmark$,$\times$,$\checkmark$,$\times$,$\checkmark$,$\times$,$\checkmark$,$\times$,$\checkmark$,$\times$,$\checkmark$,$\times$} {
			\node[minimum size=7mm,draw=black,line width=0.2pt] at (2,-\i) {\a};
		}
		
		\foreach \a [count=\i] in {DYN,DYN,DYN,DYN,F05,F05,F05,F05,F10,F10,F10,F10} {
			\pgfmathparse{\i+2}
			\edef\newI{\pgfmathresult}
			\node[line width=0.2pt,draw=black,inner sep=0pt,text width=7mm,minimum height=7mm,align=center] at (\newI,-13) {\footnotesize \a};
		}
		
		\foreach \a [count=\i] in {$\checkmark$,$\checkmark$,$\times$,$\times$,$\checkmark$,$\checkmark$,$\times$,$\times$,$\checkmark$,$\checkmark$,$\times$,$\times$} {
			\pgfmathparse{\i+2}
			\edef\newI{\pgfmathresult}
			\node[minimum size=7mm,draw=black,line width=0.2pt] at (\newI,-14) {\a};
		}

		\foreach \a [count=\i] in {$\checkmark$,$\times$,$\checkmark$,$\times$,$\checkmark$,$\times$,$\checkmark$,$\times$,$\checkmark$,$\times$,$\checkmark$,$\times$} {
			\pgfmathparse{\i+2}
			\edef\newI{\pgfmathresult}
			\node[minimum size=7mm,draw=black,line width=0.2pt] at (\newI,-15) {\a};
		}
		
		\node[minimum size=6mm] at (14.8,-13) {\footnotesize \textbf{IT}};
		\node[minimum size=6mm] at (14.8,-14) {\footnotesize \textbf{L}};
		\node[minimum size=6mm] at (14.8,-15) {\footnotesize \textbf{M}};

        \node[minimum size=15mm] at (14.8,-16) {\footnotesize \textbf{}};
        \node[minimum size=35mm] at (14.8,2) {\footnotesize \textbf{}};
		
		\node[minimum size=6mm] at (1.8,-12.8) {$\alpha$};
		\node[minimum size=6mm] at (2.2,-13.2) {$\beta$};
		
		\node[minimum size=6mm,path picture={\draw (path picture bounding box.south west) -- (path picture bounding box.north east);}] at (2,-13) { };

	\end{tikzpicture}
	\caption{Heatmap evaluation of the \textbf{Relevance} component of the generated questions. Rows represent configurations $\alpha$ and columns represent configurations $\beta$. Each entry shows the score $\gamma(\alpha,\beta)$. 
		\textbf{Legend:} \emph{IT}: Iteration Type; \emph{L}: Student Level; \emph{M}: Presence of Supporting Materials; \emph{DYN}: Dynamic number of iterations; \emph{F05}: Fixed to 5 iterations; \emph{F10}: Fixed to 10 iterations.}
	
	\label{HM:relevance}
\end{figure}

\begin{figure}
	\centering
	\begin{tikzpicture}[scale=0.85]
		
		\node[font=\large\bfseries] at (7.5,0.2) {Depth Evaluation};
		
		\foreach \y [count=\n] in {
			{-1, 0.00, 0.32, 0.00, 0.13, 0.00, 0.73, 0.00, 0.73, 0.08, 0.85, 0.04},
            {1.00, -1, 0.95, 0.62, 0.65, 0.34, 1.00, 0.14, 0.96, 0.23, 1.00, 0.14},
            {0.68, 0.05, -1, 0.09, 0.28, 0.07, 0.89, 0.05, 0.83, 0.21, 0.88, 0.04},
            {1.00, 0.38, 0.91, -1, 0.64, 0.25, 1.00, 0.00, 0.96, 0.18, 1.00, 0.08},
            {0.87, 0.35, 0.72, 0.36, -1, 0.34, 0.98, 0.12, 0.89, 0.10, 0.96, 0.09},
            {1.00, 0.66, 0.93, 0.75, 0.66, -1, 0.98, 0.13, 0.98, 0.20, 1.00, 0.11},
            {0.27, 0.00, 0.11, 0.00, 0.02, 0.02, -1, 0.00, 0.63, 0.00, 0.78, 0.02},
            {1.00, 0.86, 0.95, 1.00, 0.88, 0.87, 1.00, -1, 0.94, 0.22, 1.00, 0.36},
            {0.27, 0.04, 0.17, 0.04, 0.11, 0.02, 0.37, 0.06, -1, 0.04, 0.47, 0.00},
            {0.92, 0.77, 0.79, 0.82, 0.90, 0.80, 1.00, 0.78, 0.96, -1, 1.00, 0.58},
            {0.15, 0.00, 0.12, 0.00, 0.04, 0.00, 0.22, 0.00, 0.53, 0.00, -1, 0.00},
            {0.96, 0.86, 0.96, 0.92, 0.91, 0.89, 0.98, 0.64, 1.00, 0.42, 1.00, -1},
		} {
			
			\foreach \x [count=\m] in \y {
				
				\ifnum\m=\n
				\node[minimum size=8mm,draw=black,line width=0.1pt,fill=white] at (\m+2,-\n) {$\bullet$}; 
				\else
				
				\pgfmathparse{\x<0.50 ? "black" : "white"}
				\edef\textcol{\pgfmathresult}
				
				\pgfmathparse{\x*100}
				\edef\mix{\pgfmathresult}
				
				\pgfmathparse{\m+2}
				\edef\newM{\pgfmathresult}
				
				\node[fill=black!\mix!white,minimum size=8mm,line width=0.1pt,draw=black,text=\textcol] at (\newM,-\n) {\footnotesize \x};
				
				\fi
			}
			
		}
		
		\node[minimum size=6mm] at (0,-0.3) {\footnotesize \textbf{IT}};
		\node[minimum size=6mm] at (1,-0.3) {\footnotesize \textbf{L}};
		\node[minimum size=6mm] at (2,-0.3) {\footnotesize \textbf{M}};

		\foreach \a [count=\i] in {DYN,DYN,DYN,DYN,F05,F05,F05,F05,F10,F10,F10,F10} {
			\node[line width=0.2pt,draw=black,inner sep=0pt,text width=7mm,minimum height=7mm,align=center] at (0,-\i) {\footnotesize \a};
		}
		
		\foreach \a [count=\i] in {$\checkmark$,$\checkmark$,$\times$,$\times$,$\checkmark$,$\checkmark$,$\times$,$\times$,$\checkmark$,$\checkmark$,$\times$,$\times$} {
			\node[minimum size=7mm,draw=black,line width=0.2pt] at (1,-\i) {\a};
		}
		
		\foreach \a [count=\i] in {$\checkmark$,$\times$,$\checkmark$,$\times$,$\checkmark$,$\times$,$\checkmark$,$\times$,$\checkmark$,$\times$,$\checkmark$,$\times$} {
			\node[minimum size=7mm,draw=black,line width=0.2pt] at (2,-\i) {\a};
		}
		
		\foreach \a [count=\i] in {DYN,DYN,DYN,DYN,F05,F05,F05,F05,F10,F10,F10,F10} {
			\pgfmathparse{\i+2}
			\edef\newI{\pgfmathresult}
			\node[line width=0.2pt,draw=black,inner sep=0pt,text width=7mm,minimum height=7mm,align=center] at (\newI,-13) {\footnotesize \a};
		}
		
		\foreach \a [count=\i] in {$\checkmark$,$\checkmark$,$\times$,$\times$,$\checkmark$,$\checkmark$,$\times$,$\times$,$\checkmark$,$\checkmark$,$\times$,$\times$} {
			\pgfmathparse{\i+2}
			\edef\newI{\pgfmathresult}
			\node[minimum size=7mm,draw=black,line width=0.2pt] at (\newI,-14) {\a};
		}

		\foreach \a [count=\i] in {$\checkmark$,$\times$,$\checkmark$,$\times$,$\checkmark$,$\times$,$\checkmark$,$\times$,$\checkmark$,$\times$,$\checkmark$,$\times$} {
			\pgfmathparse{\i+2}
			\edef\newI{\pgfmathresult}
			\node[minimum size=7mm,draw=black,line width=0.2pt] at (\newI,-15) {\a};
		}
		
		\node[minimum size=6mm] at (14.8,-13) {\footnotesize \textbf{IT}};
		\node[minimum size=6mm] at (14.8,-14) {\footnotesize \textbf{L}};
		\node[minimum size=6mm] at (14.8,-15) {\footnotesize \textbf{M}};
		
		\node[minimum size=6mm] at (1.8,-12.8) {$\alpha$};
		\node[minimum size=6mm] at (2.2,-13.2) {$\beta$};
		
		\node[minimum size=6mm,path picture={\draw (path picture bounding box.south west) -- (path picture bounding box.north east);}] at (2,-13) { };
		
		       \node[minimum size=15mm] at (14.8,-16) {\footnotesize \textbf{}};
        \node[minimum size=35mm] at (14.8,2) {\footnotesize \textbf{}};
		
	\end{tikzpicture}
	\caption{Heatmap evaluation of the \textbf{Depth} component of the generated questions. Rows represent configurations $\alpha$ and columns represent configurations $\beta$. Each entry shows the score $\gamma(\alpha,\beta)$. 
		\textbf{Legend:} \emph{IT}: Iteration Type; \emph{L}: Student Level; \emph{M}: Presence of Supporting Materials; \emph{DYN}: Dynamic number of iterations; \emph{F05}: Fixed to 5 iterations; \emph{F10}: Fixed to 10 iterations.}
	
	\label{HM:depth}
\end{figure}

\begin{figure}
	\centering
	\begin{tikzpicture}[scale=0.85]
	
		\node[font=\large\bfseries] at (7.5,0.2) {Overall Quality};
	
		\foreach \y [count=\n] in {
			{-1, 0.03, 0.40, 0.03, 0.26, 0.18, 0.98, 0.10, 0.91, 0.85, 0.91, 0.55},
            {0.97, -1, 0.95, 0.80, 0.56, 0.66, 1.00, 0.69, 1.00, 1.00, 1.00, 1.00},
            {0.60, 0.05, -1, 0.22, 0.53, 0.32, 0.98, 0.40, 0.96, 0.76, 1.00, 0.83},
            {0.97, 0.20, 0.78, -1, 0.62, 0.63, 1.00, 0.44, 1.00, 0.95, 1.00, 0.92},
            {0.74, 0.44, 0.47, 0.38, -1, 0.46, 0.95, 0.38, 0.89, 0.45, 0.96, 0.55},
            {0.82, 0.34, 0.68, 0.37, 0.54, -1, 0.98, 0.42, 1.00, 0.79, 1.00, 0.59},
            {0.02, 0.00, 0.02, 0.00, 0.05, 0.02, -1, 0.02, 0.61, 0.14, 0.77, 0.02},
            {0.90, 0.31, 0.60, 0.56, 0.62, 0.58, 0.98, -1, 0.91, 0.63, 1.00, 0.62},
            {0.09, 0.00, 0.04, 0.00, 0.11, 0.00, 0.39, 0.09, -1, 0.17, 0.58, 0.21},
            {0.15, 0.00, 0.24, 0.05, 0.55, 0.21, 0.86, 0.37, 0.83, -1, 0.96, 0.54},
            {0.09, 0.00, 0.00, 0.00, 0.04, 0.00, 0.23, 0.00, 0.42, 0.04, -1, 0.04},
            {0.45, 0.00, 0.17, 0.08, 0.45, 0.41, 0.98, 0.38, 0.79, 0.46, 0.96, -1},
		} {
	
			\foreach \x [count=\m] in \y {
				
				\ifnum\m=\n
					\node[minimum size=8mm,draw=black,line width=0.1pt,fill=white] at (\m+2,-\n) {$\bullet$}; 
				\else
				
					\pgfmathparse{\x<0.50 ? "black" : "white"}
					\edef\textcol{\pgfmathresult}
					
					\pgfmathparse{\x*100}
					\edef\mix{\pgfmathresult}
					
					\pgfmathparse{\m+2}
					\edef\newM{\pgfmathresult}
					
					\node[fill=black!\mix!white,minimum size=8mm,line width=0.1pt,draw=black,text=\textcol] at (\newM,-\n) {\footnotesize \x};
				
				\fi
			}
	
	    }
	
	    \node[minimum size=6mm] at (0,-0.3) {\footnotesize \textbf{IT}};
	    \node[minimum size=6mm] at (1,-0.3) {\footnotesize \textbf{L}};
	    \node[minimum size=6mm] at (2,-0.3) {\footnotesize \textbf{M}};

           \node[minimum size=15mm] at (14.8,-16) {\footnotesize \textbf{}};
        \node[minimum size=35mm] at (14.8,2) {\footnotesize \textbf{}};
	
		\foreach \a [count=\i] in {DYN,DYN,DYN,DYN,F05,F05,F05,F05,F10,F10,F10,F10} {
		    \node[line width=0.2pt,draw=black,inner sep=0pt,text width=7mm,minimum height=7mm,align=center] at (0,-\i) {\footnotesize \a};
		}
	  
		\foreach \a [count=\i] in {$\checkmark$,$\checkmark$,$\times$,$\times$,$\checkmark$,$\checkmark$,$\times$,$\times$,$\checkmark$,$\checkmark$,$\times$,$\times$} {
		  	\node[minimum size=7mm,draw=black,line width=0.2pt] at (1,-\i) {\a};
		}
	  
		\foreach \a [count=\i] in {$\checkmark$,$\times$,$\checkmark$,$\times$,$\checkmark$,$\times$,$\checkmark$,$\times$,$\checkmark$,$\times$,$\checkmark$,$\times$} {
		  	\node[minimum size=7mm,draw=black,line width=0.2pt] at (2,-\i) {\a};
		}
	  
		\foreach \a [count=\i] in {DYN,DYN,DYN,DYN,F05,F05,F05,F05,F10,F10,F10,F10} {
		   	\pgfmathparse{\i+2}
		   	\edef\newI{\pgfmathresult}
		  	\node[line width=0.2pt,draw=black,inner sep=0pt,text width=7mm,minimum height=7mm,align=center] at (\newI,-13) {\footnotesize \a};
	 	}
	  
		\foreach \a [count=\i] in {$\checkmark$,$\checkmark$,$\times$,$\times$,$\checkmark$,$\checkmark$,$\times$,$\times$,$\checkmark$,$\checkmark$,$\times$,$\times$} {
			\pgfmathparse{\i+2}
		  	\edef\newI{\pgfmathresult}
		  	\node[minimum size=7mm,draw=black,line width=0.2pt] at (\newI,-14) {\a};
		}

		\foreach \a [count=\i] in {$\checkmark$,$\times$,$\checkmark$,$\times$,$\checkmark$,$\times$,$\checkmark$,$\times$,$\checkmark$,$\times$,$\checkmark$,$\times$} {
		  	\pgfmathparse{\i+2}
		  	\edef\newI{\pgfmathresult}
		  	\node[minimum size=7mm,draw=black,line width=0.2pt] at (\newI,-15) {\a};
		}
	  
		\node[minimum size=6mm] at (14.8,-13) {\footnotesize \textbf{IT}};
		\node[minimum size=6mm] at (14.8,-14) {\footnotesize \textbf{L}};
		\node[minimum size=6mm] at (14.8,-15) {\footnotesize \textbf{M}};
		
		\node[minimum size=6mm] at (1.8,-12.8) {$\alpha$};
		\node[minimum size=6mm] at (2.2,-13.2) {$\beta$};
		
		\node[minimum size=6mm,path picture={\draw (path picture bounding box.south west) -- (path picture bounding box.north east);}] at (2,-13) { };

	\end{tikzpicture}
	\caption{Heatmap evaluation of the \textbf{Overall Quality} of the generated questions. Rows represent configurations $\alpha$ and columns represent configurations $\beta$. Each entry shows the score $\gamma(\alpha,\beta)$. 
	\textbf{Legend:} \emph{IT}: Iteration Type; \emph{L}: Student Level; \emph{M}: Presence of Supporting Materials; \emph{DYN}: Dynamic number of iterations; \emph{F05}: Fixed to 5 iterations; \emph{F10}: Fixed to 10 iterations.}

	\label{HM:overall}
\end{figure}

Regarding \textbf{clarity} (see Figure~\ref{HM:clarity}), the results show a pronounced advantage for DYN-based configurations, particularly over F10. The best-performing settings involve either or both student-level specification and materials ($L\checkmark$, $M\checkmark$). This suggests that contextual grounding and adaptive iteration help maintain syntactic and semantic fluency, while prolonged refinement (as in F10) may lead to verbosity or self-contradiction.  In the \textbf{relevance} domain (Figure~\ref{HM:relevance}), dynamic iteration (DYN) configurations maintain a strong advantage across most pairwise comparisons, particularly when combined with supporting materials ($M\checkmark$), often yielding $\gamma(\alpha,\beta) > 0.9$ against fixed-length variants without materials. However, some Fixed-5 (F05) configurations with materials achieve competitive results (around $\gamma \approx 0.7$) suggesting that contextual grounding can partially offset the benefits of adaptivity. In contrast, F10 settings without materials ($M\times$) consistently perform worst, with most $\gamma$ values below 0.2 when compared to DYN alternatives. These trends indicate that both adaptivity and contextual information play complementary roles in ensuring that the generated questions remain topically aligned with the provided concepts and instructional objectives. For \textbf{depth} (Figure~\ref{HM:depth}), the pattern is more nuanced. DYN with materials remains dominant, but certain F10 configurations approach parity. This suggests that deeper reflection may sometimes benefit from longer, multi-turn refinement, provided clarity is not degraded. Nevertheless, DYN–$M\checkmark$ settings achieve the highest $\gamma$ values overall, indicating that adaptive strategies more reliably yield thought-provoking questions with fewer steps.

Dynamic iteration (DYN) schemes consistently outperform both fixed-length settings in Overall Quality (Figure~\ref{HM:overall}). Configurations combining DYN with provided materials ($M\checkmark$) dominate nearly all comparisons, often achieving $\gamma(\alpha,\beta) > 0.9$ against fixed alternatives. This indicates that adaptively terminating the dialogue when the Teacher--Educator emits the ``Great question!'' signal produces higher-quality reflective prompts than enforcing a fixed number of refinement rounds. Including the student level ($L\checkmark$) further improves quality, although its effect is secondary to the iteration regime. Regarding RQ1, these results suggest that interaction design (in particular, dynamic stopping) is the primary driver of quality gains, with contextual information (student level and materials) providing complementary but smaller benefits.

Below we summarize a few key takeaways from RQ1 for educators:

\begin{itemize}
    \item \textbf{Adaptive interaction matters:} Allowing the model to terminate adaptively (DYN) produces consistently better reflective questions than enforcing fixed-length refinement;
    \item \textbf{Contextual grounding enhances outcomes:} Providing supporting materials ($M\checkmark$) and student-level information ($L\checkmark$) improves clarity and relevance, enabling more targeted and meaningful question generation;
    \item \textbf{Balance depth with coherence:} While longer dialogues can foster deeper reflection, adaptive termination prevents redundancy and maintains fluency, which is crucial for high-quality pedagogical output.
\end{itemize}

\subsection{RQ2: Protocol Effectiveness}

For RQ2, we again use the normalized preference index $\gamma(\alpha,\beta)$ defined in Section~\ref{subsec:rq1-design}, but now fix $\alpha$ to the \emph{reflection-in-reflection} configuration with dynamic stopping (DYN) and $\beta$ to the one-shot baseline using the same backbone model under the same contextual condition (presence/absence of student level $L$ and materials $M$). Thus, each cell in Figure~\ref{HM:another} reports $\gamma(\alpha,\beta)\in[0,1]$, where values above $0.5$ indicate that the evaluator more often prefers questions from the reflection-in-reflection protocol over those from the one-shot baseline, and values below $0.5$ indicate the opposite.

Across all context settings, reflection-in-reflection yields a clear advantage in \textbf{relevance}: $\gamma$ ranges from $0.60$ to $0.92$, indicating that the multi-turn, Socratic protocol more reliably keeps questions aligned with the topic and key concepts than a single-shot prompt, even when no student level or materials are provided (row $L\times,M\times$: $\gamma_{\text{rel}}=0.92$). \textbf{Depth} also generally benefits from the two-agent design: the average depth score is above $0.6$, with particularly strong gains when only the level is specified or when no contextual signals are provided (e.g., $L\checkmark,M\times$: $\gamma_{\text{depth}}=0.77$; $L\times,M\times$: $\gamma_{\text{depth}}=0.91$). This suggests that the iterative Socratic feedback loop is especially helpful for moving beyond surface-level prompts when the initial context is underspecified. The effects on \textbf{clarity} and \textbf{overall quality} are more moderate but still positive on average. Clarity scores hover around or slightly above $0.5$ in three out of four conditions (0.64, 0.60, 0.60), indicating a small but consistent edge for reflection-in-reflection when either level or materials are available. The exception is the no-context case ($L\times,M\times$), where the one-shot baseline produces simpler, more fluent questions ($\gamma_{\text{clar}}=0.20$), while reflection-in-reflection trades some linguistic polish for greater conceptual relevance and depth. Overall quality behaves similarly: $\gamma$ is close to or slightly above $0.5$ in all settings (0.58, 0.54, 0.50, 0.60), with the largest margin again in the no-context condition, where the two-agent protocol appears to compensate for missing scaffolding by constructing globally stronger prompts than the baseline.

\begin{figure}
	\centering
	\begin{tikzpicture}[scale=0.85]
	
		\node[font=\large\bfseries] at (2.5,0.2) {R in R vs One-shot LLM Prompt};
	
		\foreach \y [count=\n] in {
		{0.64, 0.75, 0.46, 0.58},
            {0.60, 0.67, 0.77, 0.54},
            {0.60, 0.60, 0.36, 0.50},
            {0.20, 0.92, 0.91, 0.60},
		} {
	
			\foreach \x [count=\m] in \y {
				
				\ifnum\m=1000
					\node[minimum size=8mm,draw=black,line width=0.1pt,fill=white] at (\m+2,-\n) {$\bullet$}; 
				\else
				
					\pgfmathparse{\x<0.50 ? "black" : "white"}
					\edef\textcol{\pgfmathresult}
					
					\pgfmathparse{\x*100}
					\edef\mix{\pgfmathresult}
					
					\pgfmathparse{\m+1}
					\edef\newM{\pgfmathresult}
					
					\node[fill=black!\mix!white,minimum size=8mm,line width=0.1pt,draw=black,text=\textcol] at (\newM,-\n) {\footnotesize \x};
				
				\fi
			}
	
	    }
	
	    \node[minimum size=6mm] at (0,-0.3) {\footnotesize \textbf{L}};
	    \node[minimum size=6mm] at (1,-0.3) {\footnotesize \textbf{M}};

            \node[minimum size=6mm, rotate=90, transform shape] at (2,-5.5) {\textbf{Clarity}};
            \node[minimum size=6mm, rotate=90, transform shape] at (3,-5.5) {\textbf{Relevance}};
            \node[minimum size=6mm, rotate=90, transform shape] at (4,-5.5) {\textbf{Depth}};
            \node[minimum size=6mm, rotate=90, transform shape] at (5,-5.5) {\textbf{O. Quality}};

	  
		\foreach \a [count=\i] in {$\checkmark$,$\checkmark$,$\times$,$\times$} {
		  	\node[minimum size=7mm,draw=black,line width=0.2pt] at (0,-\i) {\a};
		}
	  
		\foreach \a [count=\i] in {$\checkmark$,$\times$,$\checkmark$,$\times$} {
		  	\node[minimum size=7mm,draw=black,line width=0.2pt] at (1,-\i) {\a};
		}

	\end{tikzpicture}
	\caption{Comparison between the \emph{reflection-in-reflection} (R-in-R) protocol and a one-shot LLM baseline across evaluation criteria and context settings. Rows correspond to context configurations given by the presence ($\checkmark$) or absence ($\times$) of explicit student level (\textbf{L}) and supporting materials (\textbf{M}). Columns correspond to the four evaluation criteria (clarity, relevance, depth, overall quality). Each cell shows the normalized preference index $\gamma \in [0,1]$, where values above $0.5$ indicate that questions from the R-in-R protocol are preferred over those from the one-shot baseline.}

	\label{HM:another}
\end{figure}

Taken together, these results indicate that the reflection-in-reflection protocol is indeed judged superior to a one-shot LLM baseline for RQ2, especially in terms of aligning questions with the intended topic and concepts and promoting deeper reflection. The gains are most pronounced when some contextual information (student level and/or materials) is supplied, but even in the weakest setting ($L\times,M\times$), the two-agent, multi-turn architecture delivers substantially higher relevance and depth, at the cost of slightly reduced surface clarity. For practitioners, this suggests that pairing a modest backbone model with a structured, Socratic interaction protocol can yield better reflective prompts than simply ``asking once'', particularly when at least minimal curricular context is provided.

\subsection{Qualitative Examples of  Dialogues}
\label{subsec:qualitative-examples}

The complete set of generated questions and full dialogue traces (including all intermediate turns of both agents) is available in our repository as supplementary material. In this subsection, we present a small number of illustrative excerpts that highlight interesting behaviours of the \emph{reflection-in-reflection} protocol: (i) high-quality questions, (ii) dialogue drift in longer interactions, and (iii) the extent to which questions can evolve across iterations. For readability, we omit the fixed role prefixes from the outputs: in the raw data, Student--Teacher turns are prefixed with \texttt{The Student's response:} and Teacher--Educator turns with \texttt{The Teacher's feedback:}, but these markers do not affect the substantive content.

\paragraph{High-quality question example.}
Example~\ref{list:ideal-question} shows a question that the evaluator rated as having high overall quality. It was generated under the dynamic stopping regime (DYN) with student-level information ($L\checkmark$) but without supporting materials ($M\times$). The first part contains the final reflection question, while the second part is the Student--Teacher's brief rationale, explaining how the latest revision responds to the Teacher--Educator's feedback. The question explicitly asks for ``specific ways'' in which decentralization impacts data transmission and makes the role of data packets salient, thereby narrowing the scope while still encouraging analytical reasoning. At the same time, the language remains accessible for 8th–9th grade students, illustrating how the protocol can balance depth with age-appropriate phrasing.

\newpage

\begin{lstlisting}[
  caption={Sample LLM-generated reflection question and rationale.},
  label=list:ideal-question,
  captionpos=t,
  basicstyle=\small\ttfamily,
  backgroundcolor=\color{backgroundgray},
  breaklines=true,
  breakatwhitespace=true,
  breakindent=0pt
]
In what specific ways does decentralization impact the process of data transmission over the internet from servers to clients, and how do data packets facilitate that process?

In this revision, I replaced "influence" with "impact" and explicitly invited students to think about "specific ways" decentralization affects data transmission. This clarification helps to narrow their focus on tangible aspects of how decentralization operates in relation to the flow of data. By highlighting the role of data packets in facilitating this process, I encourage students to better understand the intricate details of internet functionality. Overall, this question promotes a more targeted exploration of the topic while remaining accessible for 8th and 9th graders.
\end{lstlisting}

\paragraph{Dialogue drift in longer interactions.}
As can be inferred from the quantitative results in RQ1, longer dialogues are more prone to drifting away from the core task of question refinement (particularly in the F10 condition, which is consistently evaluated with quite low relevance - see Figure~\ref{HM:relevance}). Example~\ref{list:off-topic-question} illustrates a Student--Teacher response that largely abandons question revision and instead elaborates a mini lesson plan on how to facilitate classroom discussion. This behaviour is not solely attributable to the Student--Teacher: Example~\ref{list:off-topic-question-teacher} shows the corresponding Teacher--Educator turn, which explicitly asks about strategies for ``ensuring every student's voice is heard'' and ``encourag[ing] them to build on each other's ideas''. Such prompts shift the focus from question design to classroom orchestration, thereby steering the dialogue outside the intended scope of the system.
.

\begin{lstlisting}[
  caption={Student--Teacher response drifting off-topic.},
  label=list:off-topic-question,
  captionpos=t,
  basicstyle=\small\ttfamily,
  backgroundcolor=\color{backgroundgray},
  breaklines=true,
  breakatwhitespace=true,
  breakindent=0pt,
  escapechar=|,
]
To facilitate the discussion effectively and ensure every student's voice is heard, I plan to implement the following strategies:

1. |\textbf{Think-Pair-Share}|: I will first ask students to reflect individually on the questions before discussing their thoughts with a partner. This gives quieter students a chance to articulate their ideas in a smaller setting before sharing with the larger group.

2. |\textbf{Structured~Turn-Taking}|: I will establish a system where each student has an opportunity to speak, perhaps by passing a visible object (like a talking stick) to ensure that everyone gets a turn and encourages those who might hesitate to share.

3. |\textbf{Building~on~Ideas}|: I will prompt students to respond to each other's thoughts by asking follow-up questions, such as, ``Who agrees with this point, and can you add to it?'' or ``Does anyone have a different perspective to share?''

By employing these strategies, I aim to create an inclusive and balanced conversation, promoting a supportive environment where all students feel empowered to express their views and engage with their peers' ideas.
\end{lstlisting}

\begin{lstlisting}[
  caption={Teacher--Educator question that steers the dialogue off-topic (iteration 7 in attempt 1 in fixed10iter.json).},
  label=list:off-topic-question-teacher,
  captionpos=t,
  basicstyle=\small\ttfamily,
  backgroundcolor=\color{backgroundgray},
  breaklines=true,
  breakatwhitespace=true,
  breakindent=0pt
]
How do you plan to facilitate the discussion to ensure every student's voice is heard? Might including strategies for promoting quieter students to share their thoughts help create a more balanced conversation? Additionally, how could you encourage them to build on each other's ideas during the discussion?
\end{lstlisting}

\paragraph{Evolution of a question across iterations.}
Finally, Examples~\ref{list:0-iter-question} and~\ref{list:10-iter-question} illustrate how much a single question can change over the course of a dialogue. Example~\ref{list:0-iter-question} shows the initial question proposed by the Student--Teacher before any feedback (iteration 0), whereas Example~\ref{list:10-iter-question} shows the same question after ten refinement steps (F10 condition). Explanations are omitted for brevity.

The initial question is concise and reasonably well-formed, focusing on a single relationship between decentralization and packet travel. After ten iterations, the question has evolved into a multi-part prompt that asks students to identify problems, propose solutions, discuss limitations, and connect their reasoning to real-world examples. In our best-iteration analysis, the final version is judged superior in terms of \emph{depth}, but earlier iterations sometimes achieve higher \emph{clarity} or \emph{relevance}. This mirrors the quantitative patterns from RQ1: additional turns can enrich cognitive demand and reflective scope, but beyond a certain point may introduce complexity that not all learners will find accessible.

\begin{lstlisting}[
  caption={Reflection question after 0 dialog iterations.},
  label=list:0-iter-question,
  captionpos=t,
  basicstyle=\small\ttfamily,
  backgroundcolor=\color{backgroundgray},
  breaklines=true,
  breakatwhitespace=true,
  breakindent=0pt
]
How does the decentralization of the internet impact the way data packets travel from a server to a client device?

[...]
\end{lstlisting}

\begin{lstlisting}[
  caption={Reflection question after 10 iterations.},
  label=list:10-iter-question,
  captionpos=t,
  basicstyle=\small\ttfamily,
  backgroundcolor=\color{backgroundgray},
  breaklines=true,
  breakatwhitespace=true,
  breakindent=0pt
]
What are some different opinions on how servers, routers, and clients in a decentralized internet affect the speed and reliability of data packets? First, identify problems related to data congestion and redundancy. Next, suggest solutions to these problems, and discuss the challenges or limitations of implementing your solutions. Finally, how might these solutions impact users' experiences with the internet? Please provide real-world examples to support your points.

[...]
\end{lstlisting}

\section{Conclusion}
\label{sec:conclusion}

This paper addressed the challenge of generating high-quality reflection questions for teaching, a task that is pedagogically important but time-consuming and unevenly supported across educators. We proposed a \emph{reflection-in-reflection} framework in which two LLM-based agents (a Student--Teacher and a Teacher--Educator) jointly construct a single reflection question through a Socratic, multi-turn dialogue. The Teacher--Educator provides targeted coaching questions and dynamically decides when the question is satisfactory, rather than enforcing a fixed number of refinement steps.

Our experiments, based on an LLM evaluator and a pairwise comparison protocol, yielded two main findings. First (RQ1), interaction design matters: dynamic stopping (DYN) consistently outperforms fixed-length regimes (F05, F10) across clarity, relevance, depth, and overall quality. The best results are obtained when DYN is combined with supporting materials and explicit student-level information, indicating that adaptivity and contextual grounding are complementary. Second (RQ2), the reflection-in-reflection protocol is generally preferred over a one-shot baseline using the same backbone model, with the clearest gains in relevance and depth, particularly when at least minimal context (student level and/or materials) is available.

In addition to its experimental contributions, the reflection-in-reflection framework offers direct implications for teacher education and pedagogical practice. For instance, within teacher education programs, the system could be implemented as a structured micro-practice tool, allowing preservice teachers to practice generating, refining, and evaluating reflection questions while receiving immediate coaching from the Teacher–Educator agent. Such integration may facilitate the internalization of effective questioning strategies and foster a deeper metacognitive understanding of the pedagogical role of reflection questions.

In daily teaching practice, the framework can function as a planning aid that facilitates the rapid development of contextualized reflection prompts tailored to specific lesson content, learner characteristics, and assessment data. This support may help teachers sustain reflective rigor despite time constraints. Additionally, schools and districts could incorporate the tool into professional learning communities, enabling collaborative evaluation of AI-generated prompts and fostering collective inquiry into instructional practice.
The system could also be adapted for student-facing platforms, enabling learners to collaboratively construct reflection questions with targeted scaffolding. This approach has the potential to enhance self-regulated learning and reflective capacity. Collectively, these application pathways underscore the practical significance of adaptive, dialogic large language model frameworks in advancing reflective practice throughout the teacher learning ecosystem.

This study has two main limitations. Most notably, evaluation is conducted entirely with an LLM-based judge rather than human raters. This choice is driven by the combinatorial size of the experiment and the complexity of the pairwise protocol, as it would be extremely time-consuming and cognitively taxing (and likely quite tedious) to ask teachers or students to perform all comparisons. At the same time, our framework can serve as a screening tool: the LLM-based evaluation narrows down the space of configurations and identifies a small set of promising profiles that are most worth presenting to humans. Future work should therefore complement our results with classroom studies in which teachers and students assess these selected configurations directly. In addition, we focus on relatively lightweight backbone models; comparing our protocol against more advanced reasoning-oriented LLMs is an important next step, though potentially costly given the large number of generations and evaluations involved. Future work will (most likely) therefore explore hybrid schemes that combine our interaction design with stronger reasoning models in a cost-aware way, and mixed human–LLM evaluation setups that could help to capture classroom usefulness while retaining scalability.


\section*{Acknowledgements}
The authors thank AI~dětem for supporting this research with OpenAI credits and for providing the instructional materials used in the experiments.

\section*{Data}

The data generated in this study are available in our public repository at
\url{https://github.com/Ender00420/Reflecting-in-the-Reflection}.
They may be freely used for research purposes, provided that this paper is cited.

\bibliographystyle{unsrtnat}
\bibliography{arxiv_bib}  






\end{document}